# Shadow-Background-Noise 3D Spatial Decomposition Using Sparse Low-Rank Gaussian Properties for Video-SAR Moving Target Shadow Enhancement

Xiaowo Xu, Xiaoling Zhang, Tianwen Zhang, Zhenyu Yang, Jun Shi, and Xu Zhan

*Abstract*—Moving target shadows among video synthetic aperture radar (Video-SAR) images are always interfered by low scattering backgrounds and cluttered noises, causing poor detection-tracking accuracy. Thus, a shadow-background-noise 3D spatial decomposition (SBN-3D-SD) model is proposed to enhance shadows for higher detection-tracking accuracy. It leverages the sparse property of shadows, the low-rank property of backgrounds, and the Gaussian property of noises to perform 3D spatial three-decomposition. It separates shadows from backgrounds and noises by the alternating direction method of multipliers (ADMM). Results on the Sandia National Laboratories (SNL) data verify its effectiveness. It boosts the shadow saliency from the qualitative and quantitative evaluation. It boosts the shadow detection accuracy of Faster R-CNN, RetinaNet and YOLOv3. It also boosts the shadow tracking accuracy of TransTrack, FairMOT and ByteTrack.

## I. INTRODUCTION

VIDEO synthetic aperture radar (Video-SAR) [1] can monitor targets continuously, widely-used in the ground vehicle surveillance. Since Sandia National Laboratories (SNL) invented Video-SAR in 2003, it has attracted much attention by many scholars [2]–[7] in the ground moving target monitoring.

Video-SAR moving target imaging has azimuth offsets and defocus, but shadows left by targets can reflect its real position. Thus, the shadow detection-tracking is a more effective way to monitor moving targets. Zhao *et al.* [2] adopted the spatial-temporal information to track shadows. Xu *et al.* [3] used geometric relationships between moving targets and shadows in position and size for ground moving target indication (GMTI). Bao *et al.* [4] proposed a network to detect shadows. Zhong *et al.* [5] used shadow and echo energy to reduce false alarms and missed-detections. Tian *et al.* [6] proposed an expanding-shrinking strategy-based particle filter for the detection-tracking of moving-target shadows. Ding *et al.* [7] applied Faster R-CNN to detect shadows. Yet, these methods did not deeply analyze differences between shadows and backgrounds-noises, so shadows are still submerged by low-scattering backgrounds and cluttered noises, hindering further detection-tracking accuracy improvements.

Thus, a shadow-background-noise 3D spatial decomposition (SBN-3D-SD) model is proposed to enhance shadows. We discover and analyze the sparse property of shadows, the low-rank property of backgrounds, and the Gaussian property of noises. A 3D-tensor spatial three-decomposition model is established

This work was supported by the National Natural Science Foundation of China under Grant 61571099. *(Corresponding author: Xiaoling Zhang.)*

The authors are with the School of Information and Communication Engineering, University of Electronic Science and Technology of China, Chengdu, China. (e-mail: xuxiaowo@std.uestc.edu.cn, xlzhang@uestc.edu.cn, twzhang@std.uestc.edu.cn, zhenyuy@std.uestc.edu.cn, shijun@uestc.edu.cn, zhanxu@std.uestc.edu.cn)

to separate shadows from backgrounds and noises. The alternating direction method of multipliers (ADMM) [8] is used to solve the model. We perform experiments on the SNL data [1] to verify SBN-3D-SD's effectiveness. Results reveal that shadows are enhanced from the qualitative and quantitative evaluation. For this advantage, SBN-3D-SD improves the shadow detection accuracy of Faster R-CNN [9], RetinaNet [10], and YOLOv3 [11], and improves the shadow tracking performance of TransTrack [12], FairMOT [13], and ByteTrack [15].

The main contributions of this letter are as follows.
1) We point out the sparse, low-rank, and Gaussian properties of the shadow, background and noise in Video-SAR data. To the best of our knowledge, this is the first time in Video-SAR moving target detection-tracking community.
2) We propose SBN-3D-SD to enhance shadows. It takes advantage of sparse, low-rank, Gaussian properties of shadows, backgrounds and noises. Shadow contrast ratio is tripled. To the best of our knowledge, this is the first time in Video-SAR moving target detection-tracking community.
3) ADMM is used to solve SBN-3D-SD. Shadows are separated from backgrounds and noises. To the best of our knowledge, this is the first attempt in Video-SAR moving target detection-tracking community.
4) SBN-3D-SD boosts the shadow detection-tracking accuracy, this is, a ~8% average precision (AP) increment and a ~10% multiple object tracking accuracy (MOTA) increment.

## II. METHODOLOGY

The continuous Video-SAR data **D** can be described as

$$\mathbf{D} = \{\mathbf{F}_1, \mathbf{F}_2, ..., \mathbf{F}_f\} \subset \mathbb{R}^{N_a \times N_r \times f} \quad (1)$$

where $\mathbf{F}_f$ is the $f$-th frame, $N_a$ is the height and $N_r$ is the width. Each frame contains three types of information components,

$$\mathbf{F}_f = \mathbf{S}_f + \mathbf{B}_f + \mathbf{N}_f \quad (2)$$

where $\mathbf{S}_f$ is the shadow, $\mathbf{B}_f$ is the background and $\mathbf{N}_f$ is the noise. With this, **D** can be represented as

$$\mathbf{D} = \mathbf{S} + \mathbf{B} + \mathbf{N} \quad (3)$$

SBN-3D-SD is responsible for separating **S** from **B** and **N** based on their differences, so their differences are the core.

### A. Property Analysis

*1) Sparse Shadows.* In Video-SAR data, the number of shadows is slim, and the scale of shadows is small. Shadows occupy a few pixels in images. A marked region in Fig. 1(a) is a moving target shadow. Fig. 1(b) shows the pixel statistics of the whole image. From Fig. 1(b), the shadow is dim whose pixel value is 29, so it is challenging to detect and track it. The shadow scale is small, because it is near the leftmost end of the curve. **S** in account for a small proportion among all pixels **D**, i.e., 0.22% as shown in Fig. 1(c), so shadows **S** have the sparse property in a single frame. Especially for the whole 3D Video-SAR data **D**, such sparse property will become more obvious.

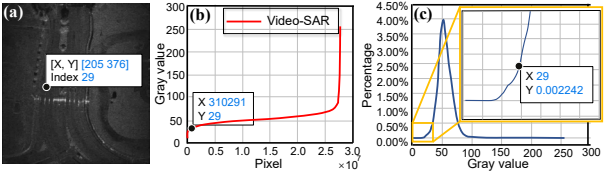

Fig. 1. Sparse shadows. (a) Shadow. (b) Pixel statistic. (c) Gray percentage.

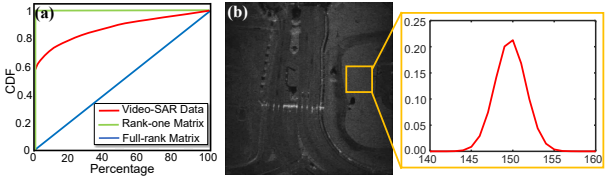

Fig. 2. (a) Low-rank backgrounds reflected by CDF. (b) Gaussian noises.

*2) Low-Rank Backgrounds.* When a region of interest is monitored by Video-SAR, the background is static and the target is moving, if multiple frames have been already registered. Thus, the static background in multiple frames is almost the same, resulting in that the information in the background is highly similar and correlated. From the entropy theory, static deterministic events (backgrounds) have low information contents, while dynamic uncertain events (moving target shadows) have high information contents. From the perspective of the matrix **D**, the background **B** with low information contents is low-rank. The full-rank matrix has the maximum information and the rank-one matrix has minimum information [16]. In the whole 3D Video-SAR data **D**, the background **B** has the low-rank property. To fully confirm this property, the cumulative distribution function (CDF) is used to analyze the spatial structure information. CDF uses the singular value decomposition to analyze matrix's low-rank, and evaluates the low-rank through the cumulative distribution function of singular values by [16]

$$\text{CDF}(k) = \frac{\sum_{j=1}^{k \cdot \min(N_c, f)} \Lambda(j, j)}{\sum_{j=1}^{\min(N_c, f)} \Lambda(j, j)} \quad (4)$$

where $\Lambda$ means the diagonal matrix composed of singular values, $\Lambda(j, j)$ means the $j$-th largest singular value, $k$ means the top $k\%$ singular values, $N_c$ means the total pixels per frame, and $f$ denotes the total video frames. The result is shown in Fig. 2(a). Here, the whole Video-SAR data is used for CDF analysis, rather than an image slice. From Fig. 2(a), the CDF of the whole Video-SAR data is closer to the rank-one matrix. This shows that the information in the Video-SAR data is redundant because of many low-value backgrounds. The energy distribution of singular values is concentrated, rather than the uniform distribution in the full-rank matrix. The above confirms the low-rank property of the background **B**.

*3) Gaussian Noises.* Noises in Video-SAR are dynamic, so it should also be sparse. This will cause sparse shadows to be easily submerged by noises. Li *et al.* [17] found empirically that in the SAR imaging process, the noises in Video-SAR images can be sometimes modeled as an approximate Gaussian distribution. We extract a noise region to analyze the noise distribution as Fig. 2(b). From Fig. 2(b), the gaussian property prior of noises fits the fact, so this difference can be used to separate shadows from noises. More accurate distributions, e.g., Exponential or

Rayleigh, are more effective, and also bring better Contrast performance (See Section V-A). However, to highlight SBN-3D-SD's universality, we still temporarily regard noises in Video-SAR as the familiar approximate Gaussian distribution in this letter, which does not undermine SBN-3D-SD's effectiveness.

### B. SBN-3D-SD

SBN-3D-SD is summarized as

$$\min_{\mathbf{S,B,N}} \|\mathbf{S}\|_0 + \xi \cdot rank(\mathbf{B}) + \gamma \cdot \|\mathbf{N}\|_{\mathrm{F}}^2$$
$$s.t. \ \mathbf{D} = \mathbf{S} + \mathbf{B} + \mathbf{N} \quad (5)$$

where $\|\cdot\|_0$ means the $\ell_0$ norm, $\|\cdot\|_{\mathrm{F}}$ means the Frobenius norm, $rank$ means the rank and $\xi$, $\gamma$ are regularization factors to balance the sparse portion, low-rank one, and Gaussian one.

*1) Shadow Constraint.* Shadow **S** occupies a few pixels in the image, and it is seen as the sparse portion, so it can be separated by the number of non-zero terms of the constraint matrix, that is, the $\ell_0$ norm can be used to constrain it.

*2) Background Constraint.* Since background **B** is static for a long time, the components in the sequence frames have linear related elements. The background can be seen as the low-rank portion. Thus, the *rank* minimization of **B** constrains the structure of the linear subspace with the background column space.

*3) Noise Constraint.* Since noise obeys the zero-mean Gaussian distribution, the Frobenius norm of the minimization matrix can be used to constrain the Gaussian distribution noise.

### C. Solution Using ADMM

To solve (5) is a non-deterministic polynomial (NP) problem, because $rank(\mathbf{B})$ and $\|\mathbf{S}\|_0$ are nonconvex, but one can use the nuclear norm $\|\mathbf{B}\|_*$ to replace $rank(\mathbf{B})$ and use the $\ell_1$ norm to replace $\|\mathbf{S}\|_0$. In this way, the raw non-convex problem of (5) is converted into a convex optimization problem as follows:

$$\min_{\mathbf{S,B,N}} \|\mathbf{S}\|_1 + \xi \cdot \|\mathbf{B}\|_* + \gamma \cdot \|\mathbf{N}\|_{\mathrm{F}}^2$$
$$s.t. \ \mathbf{D} = \mathbf{S} + \mathbf{B} + \mathbf{N} \quad (6)$$

There are two main types of methods to solve (6), i.e., the iterative thresholding [18] and ADMM [8]. Since the optimization process of the former is not smooth and the convergence speed is slow, ADMM is selected. In ADMM, first transform (6) into the augmented Lagrange multiplier function form,

$$L(\mathbf{S,B,N,Y}) = \|\mathbf{S}\|_1 + \xi \|\Pi(\mathbf{B})\|_1 + \gamma \|\mathbf{N}\|_{\mathrm{F}}^2$$
$$+ \frac{\mu}{2} \|\mathbf{D} - \mathbf{B} - \mathbf{S} - \mathbf{N}\|_{\mathrm{F}}^2 + \langle \mathbf{Y}, \mathbf{D} - \mathbf{B} - \mathbf{S} - \mathbf{N} \rangle \quad (7)$$

where $\mathbf{Y} \in \mathbb{R}^{N_c \times f}$ means the Lagrange multiplier, $\mu$ means the penalty factor, $<\mathbf{A}, \mathbf{B}> = trace(\mathbf{A}^T, \mathbf{B})$, and $\Pi$ means the moving target confidence map, which is a linear operator that weights the entries of **B** according to their confidence of corresponding to a moving target shadow such that the most probable elements are unchanged and the least are set to zero [19].

For the $k$-th iteration, when $\mathbf{Y} = \mathbf{Y}_k$, $\mu = \mu_k$, the alternating iterative shadow matrix $\mathbf{S}_k$ is

$$\mathbf{S}_{k+1} = \underset{\mathbf{S}}{\operatorname{argmin}} L(\mathbf{B}_{k+1}, \mathbf{S}, \mathbf{N}_k, \mathbf{Y}_k, \mu_k)$$
$$= \underset{\mathbf{S}}{\operatorname{argmin}} \|\mathbf{S}\|_1 + \frac{\mu_k}{2} \left\| \mathbf{S} - \left( \mathbf{D} - \mathbf{B}_{k+1} - \mathbf{N}_k + \frac{\mathbf{Y}_k}{\mu_k} \right) \right\|_{\mathrm{F}}^2 \quad (8)$$
$$= N_{1/\mu_k} \left( \mathbf{D} - \mathbf{B}_{k+1} - \mathbf{N}_k + \frac{\mathbf{Y}_k}{\mu_k} \right)$$

where $N_\varepsilon(\mathbf{Q})$ means the near-end mapping operator of the matrix **Q**. The alternating iterative background matrix $\mathbf{B}_k$ is

$$\begin{aligned}
\mathbf{B}_{k+1} &= \operatorname*{argmin}_{\mathbf{B}} L\left(\mathbf{B}, \mathbf{S}_k, \mathbf{N}_k, \mathbf{Y}_k, \mu_k\right) \\
&= \operatorname*{argmin}_{\mathbf{B}} \xi \|\mathbf{S}\|_* + \frac{\mu_k}{2}\left\|\mathbf{B} - \left(\mathbf{D} - \mathbf{S}_k - \mathbf{N}_k + \frac{\mathbf{Y}_k}{\mu_k}\right)\right\|_F^2 \\
&= U M_{\xi/\mu_k}(W) V^{\mathrm{T}} \\
&= N_{\xi/\mu_k}\left(\mathbf{D} - \mathbf{S}_k - \mathbf{N}_k + \frac{\mathbf{Y}_k}{\mu_k}\right)
\end{aligned} \quad (9)$$

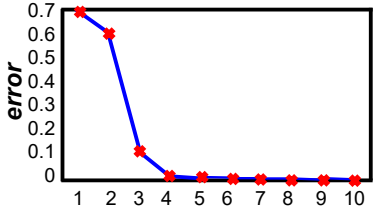

Fig. 3. The attenuation curve of the data relative transformation error.

$M_\varepsilon(\mathbf{Q}) = U N_\varepsilon(\mathbf{Q}) V^{\mathrm{T}}$ is the closed-form solution of the corresponding kernel norm optimization. $U \Sigma V^{\mathrm{T}}$ is the singular value decomposition. The alternating iterative noise matrix $\mathbf{N}_k$ is

$$\begin{aligned}
\mathbf{N}_{k+1} &= \operatorname*{argmin}_{\mathbf{N}} L\left(\mathbf{B}_{k+1}, \mathbf{S}_{k+1}, \mathbf{N}, \mathbf{Y}_k, \mu_k\right) \\
&= \operatorname*{argmin}_{\mathbf{N}} \gamma \|\mathbf{N}\|_F^2 + \frac{\mu_k}{2}\left\|\mathbf{N} - \left(\mathbf{D} - \mathbf{B}_{k+1} - \mathbf{S}_{k+1} + \frac{\mathbf{Y}_k}{\mu_k}\right)\right\|_F^2 \\
&= \left(1 + \frac{2\gamma}{\mu_k}\right)^{-1}\left(\mathbf{D} - \mathbf{B}_{k+1} - \mathbf{S}_{k+1} + \frac{\mathbf{Y}_k}{\mu_k}\right)
\end{aligned} \quad (10)$$

When $\mathbf{S} = \mathbf{S}_{k+1}$, $\mathbf{B} = \mathbf{B}_{k+1}$, and $\mathbf{N} = \mathbf{N}_{k+1}$, the Lagrange multiplier of the next iteration is

$$\mathbf{Y}_{k+1} = \mathbf{Y}_k + \mu_k\left(\mathbf{D} - \mathbf{S}_{k+1} - \mathbf{B}_{k+1} - \mathbf{N}_{k+1}\right) \quad (11)$$

, the penalty factor of the next iteration is

$$\mu_{k+1} = \rho \mu_k \quad (12)$$

where $\rho \le 1$ is a constant to reduce the learning rate.

The iteration process will be stopped when the data relative transformation *error* of two adjacent iterations is lower than a threshold 0.001. The *error* is defined by

$$error = \frac{\|\mathbf{D} - (\mathbf{B}_k + \mathbf{S}_k + \mathbf{N}_k)\|_F}{\|\mathbf{D}\|_F} \quad (13)$$

The *error* can be regarded as the relative error of Video-SAR data for each iteration. Fig. 3 shows its attenuation curve. From Fig. 3, the total iteration number is 10. Finally, the shadow $\mathbf{S}$ will be separated from the background $\mathbf{B}$ and the noise $\mathbf{N}$ by the above iteration optimization. Given a matrix $\mathbf{D} \in \mathbb{R}^{m \times n}$, for each iteration, the computational complexity to update $\mathbf{S}_k$ is $O(mn^2 + n)$, the computational complexity to update $\mathbf{B}_k$ is $O(n^3 + mn^2)$, the computational complexity to update $\mathbf{N}_k$ is $O(m^2n^2 + 2mn)$. Therefore, the time complexity of SBN-3D-BN is $O(m^2n^2 + n^3)$. In addition, its running time is 8.2478 seconds.

## III. Experiments

### A. Data

We use the public SNL Video-SAR data [1] for experiments. The data contains 900 frames. The size of each frame is $720 \times 660$. In shadow enhancement experiments, 100 frames are selected as a sub-video for shadow enhancement. In shadow detection experiments, 600 frames are selected as the training set, and 300 frames as the test set. The data of video target tracking needs to take the video sequence as the whole unit, used to ensure that the image frames in the same video sub-sequence are arranged in order. In shadow tracking experiments, six sub-videos are elected as the training set, and three ones as the test set.

### B. Details

*1) Shadow Enhancement.* We perform SBN-3D-SD on a 3D-tensor sub-video which has continuous 100 frames. The images in the sub-videos are uniformly registered by the scale-invariant feature transform (SIFT) [20]. The frame number selection criterion in a sub-video is to ensure that the moving target keeps appearing in the sub-video without leaving the observation area, which is similar to [4] and [21]. Nine sub-videos are processed by SBN-3D-SD separately. Other hyper-parameter settings are similar to [22]. The impacts of different frame numbers in a sub-video on the results will be discussed in Section V-B.

*2) Shadow Detection.* We train Faster R-CNN by 12 epochs, and train RetinaNet and YOLOv3 by 36 epochs by the stochastic gradient descent (SGD). The momentum is 0.9, the attenuation coefficient is 0.0001, and the decay rate is 0.1. The learning rate of Faster R-CNN is 0.008, and that of RetinaNet and YOLOv3 is 0.001. The batch size is set to 4. The threshold of the non-maximum suppression (NMS) is set to 0.5.

*3) Shadow Tracking.* Nine sub-videos are first preprocessed by SBN-3D-SD to enhance shadows. Images in each sub-video are inputted into different trackers (e.g., TransTrack, FairMOT, and ByteTrack) frame by frame. Here, TransTrack, FairMOT, and ByteTrack are detection-based models, so they first detect shadows in each frame, and the detection results are regarded as the initiation input of trackers so as to complete target association. We train TransTrack by 60 epochs via SGD. The initial learning rate is 0.0001, attenuation coefficient is 0.01, learning rate is attenuated at 30-epoch, and batch size is 1. We train FairMOT by 30 epochs via Adam. The initial learning rate is 0.0001, attenuation coefficient is 0.01, learning rate is attenuated at 15-epoch, and batch size is set to 4. We train ByteTrack by 50 epochs via SGD. The initial learning rate is 0.01, attenuation coefficient is 0.00001, and batch size is 2.

### C. Evaluation Criteria

*1) Shadow Enhancement.* We use the contrast, edge preservation index (EPI), and entropy to evaluate shadow quality. Contrast is defined by [23]

$$\text{Contrast} = \sum_{i,j} \delta(i,j)^2 P_\delta(i,j) \quad (14)$$

where $\delta(i,j) = |i - j|$ means the gray difference between adjacent pixels, and $P_\delta(i,j)$ means the distribution probability of pixels with gray difference of $\delta$ between adjacent pixels. The higher the contrast, the better the performance. EPI is defined by [24]

$$\text{EPI} = \frac{\sum_{i,j}\left(|I_E(i,j) - I_E(i+1,j)| + |I_E(i,j) - I_E(i,j+1)|\right)}{\sum_{i,j}\left(|I_R(i,j) - I_E(i+1,j)| + |I_R(i,j) - I_E(i,j+1)|\right)} \quad (15)$$

where $I_E(i,j)$ is the pixel value of the evaluation shadow; $I_R(i,j)$ is the pixel value of the referenced shadow. The higher EPI, the better the performance. Entropy is defined by [25]

$$\text{Entropy} = \sum_{i=0}^{255}\sum_{j=0}^{255} P_{i,j} \log P_{i,j} \quad (16)$$

where $P_{i,j}$ means the probability of combining the gray value at a pixel position with the gray distribution of surrounding pixels. Here, $P_{i,j} = f(i,j)/(W \times H)$ where $W$ and $H$ mean the image width and height, and $f(i,j)$ is the frequency of occurrence of characteristic binary $(i,j)$. Entropy reflects the information abundance of an image. Since backgrounds are suppressed, the information amount in the image is less than that in the raw image, so the smaller the entropy, the better the performance.

| Method | EPI | Entropy | Contrast |
|---|---|---|---|
| ✗ | 1.0000 | 8.8025 | 21.0104 |
| Histogram Equalization [4] | 1.3229 | 9.1268 | 29.2587 |
| Background Difference [14] | 1.7307 | 5.9007 | 38.6847 |
| SBN-3D-SD | **1.0509** | **4.9927** | **69.3711** |

| Detector | SBN-3D-SD? | TP | FP | FN | R (%) | P (%) | AP (%) |
|---|---|---|---|---|---|---|---|
| Faster R-CNN [9] | ✗ | 1167 | 584 | 535 | 68.57 | 66.65 | 60.05 |
| | ✓ | **1257** | **494** | **445** | **73.85** | **71.79** | **68.43** |
| RetinaNet [10] | ✗ | 1016 | 580 | 686 | 59.69 | 63.66 | 48.96 |
| | ✓ | **1080** | **401** | 622 | **63.45** | **72.92** | **57.42** |
| YOLOv3 [11] | ✗ | 965 | **438** | 737 | 56.70 | 68.78 | 49.75 |
| | ✓ | **1038** | 449 | 664 | **60.99** | **69.80** | **56.34** |

| Tracker | SBN-3D-SD? | FP | FN | IDSW | MOTA (%) |
|---|---|---|---|---|---|
| TransTrack [12] | ✗ | 242 | 905 | **23** | 31.26 |
| | ✓ | **158** | **786** | **23** | **43.18** |
| FairMOT [13] | ✗ | 223 | 786 | 103 | 34.67 |
| | ✓ | **154** | **699** | **90** | **44.59** |
| ByteTrack [15] | ✗ | 283 | 779 | 11 | 36.96 |
| | ✓ | **202** | **628** | **5** | **50.94** |

## IV. RESULTS

### A. Quantitative Results

*1) Shadow Enhancement.* TABLE I is the quantitative result of shadow enhancement. From TABLE I, SBN-3D-SD boosts Contrast from 21.0104 to 69.3711, increase EPI from 1.0000 to 1.0509, and decreases Entropy from 8.8025 to 4.9927, showing its superior shadow enhancement ability. SBN-3D-SD offers a higher Contrast than histogram equalization [4] and background difference [14] (i.e., SBN-3D-SD's 69.3711 > 38.6847 of background difference > 29.2587 of histogram equalization). The advantages of SBN-3D-SD are that it can concurrently consider the sparse property of shadows, the low-rank property of backgrounds, and the Gaussian property of noises. Background difference merely considers the static property of backgrounds and the dynamic property of shadows, but ignores noise properties. Moreover, background difference does not reveal deep-seated properties (i.e., the sparse property of shadows and the low-rank property of backgrounds), leading to its poor performance.

*2) Shadow Detection.* TABLE II is the quantitative result of shadow detection. From TABLE II, after shadows enhanced by SBN-3D-SD, the detection accuracies of three detectors are improved. The shadow detection AP of Faster R-CNN is enhanced by ∼8%, that of RetinaNet is enhanced by ∼9%, and that of YOLOv3 is enhanced by ∼7%. The result is logical, because obvious shadows make it easier for detectors to extract features. Finally, the detection performance is bound to become better.

*3) Shadow Tracking.* TABLE III is the quantitative result of shadow tracking. After shadows enhanced by SBN-3D-SD, the tracking accuracies of three trackers are improved. The shadow tracking MOTA of TransTrack is boosted by ∼11%, that of FairMOT is boosted by ∼10%, and that of ByteTrack is boosted by ∼14%, showing its effectiveness.

### B. Qualitative Results

*1) Shadow Enhancement.* Fig. 4 gives the qualitative shadow enhancement results. The moving target shadow is enhanced by SBN-3D-SD conspicuously from visual observation. It is better than background difference and histogram equalization methods, which shows its advantage. Fig. 5 shows three components (i.e., shadow, background, and noise) of 3D spatial decomposition using SBN-3D-SD. Fig. 5(a) is the sum of Fig. 5(b), Fig. 5(c) and Fig. 5(d). From Fig. 5, the shadow, background, and noise are successfully separated, and SBN-3D-SD has a satisfactory ability of target shadow information retention. Moreover, the fixed focus shadow enhancement (FFSE) cannot be applied to the SNL data that used in this letter. This is because the focus range system parameter of FFSE is unknown for the SNL data, so it cannot be implemented for performance comparison.

*2) Shadow Detection.* Fig. 6 shows the qualitative shadow detection results. From Fig. 6, SBN-3D-SD can avoid missed detections and suppress false alarms. This indicates that SBN-3D-SD is helpful for shadow detection.

*3) Shadow Tracking.* Fig. 7 shows the qualitative shadow tracking results. Before enhancement, there are many false alarms and missed alarms, so the trajectory interruption is very serious. After enhancement, the number of track interruptions is significantly reduced, and the number of missed alarms is also less, which is closer to the real trajectory. This indicates that SBN-3D-SD is helpful for shadow tracking.

*2) Shadow Detection.* We use the average precision (AP) to measure the performance of shadow detection, defined by [4]

$$\text{AP} = \int_0^1 P(R)dR \tag{17}$$

where $P$ is the precision, and $R$ is the recall. $P=TP/(TP+FP)$ and $R=TP/(TP+FN)$. $TP$ is the number of targets correctly identified, $FP$ is the number of non-targets incorrectly identified as targets, i.e., false alarms. $FN$ is the number of targets incorrectly classified as non-targets, i.e., missed detections.

*3) Shadow Tracking.* We use the multiple object tracking accuracy (MOTA) to measure the performance of shadow tracking, defined by [13]

$$\text{MOTA} = 1 - \frac{\sum_f (FN_f + FP_f + IDSW_f)}{\sum_f GT_f} \tag{18}$$

where $f$ is the frame number, $GT_f$ is the number of real targets at the $f$-th frame, $IDSW$ is the number of the target ID changes which is used to measure the tracking trajectory consistency.

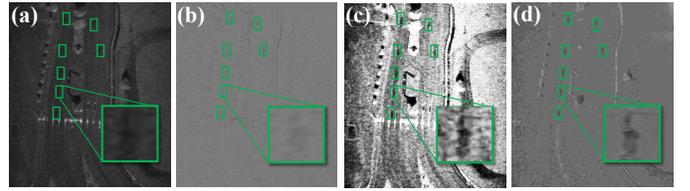

Fig. 4. (a) Shadow. (b) Shadow enhanced by background difference. (c)Shadow enhanced by histogram equalization. (d) Shadow enhanced by SBN-3D-SD.

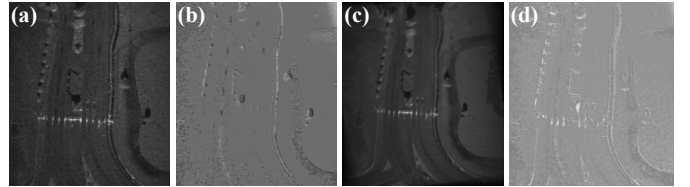

Fig. 5. Spatial decomposition results. (a) Raw image. (b) Shadow. (c) Background. (d) Noise.

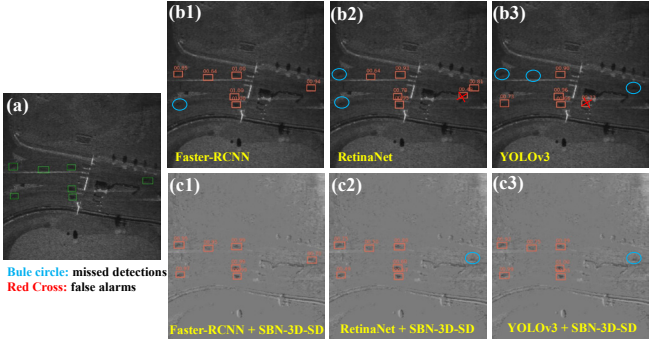

Fig. 6. Detection results. (a) GT. (b) w/o SBN-3D-SD. (c) With SBN-3D-SD.

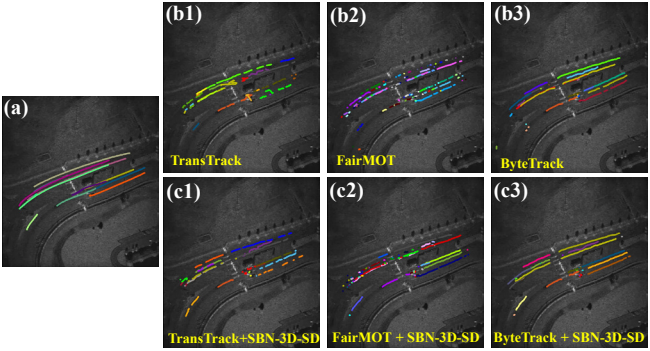

Fig. 7. Tracking results. (a) GT. (b) w/o SBN-3D-SD. (c) With SBN-3D-SD.

TABLE IV
SHADOWS QUALITY WITH DIFFERENT NOISE DISTRIBUTIONS.

| Distribution | EPI | Entropy | Contrast |
|---|---|---|---|
| -- | 1.0000 | 8.8025 | 21.0104 |
| Exponential | 0.6705 | 5.5262 | 57.9554 |
| Rayleigh | 0.7660 | 6.4322 | **69.7022** |
| Gaussian | **1.0509** | **4.9927** | 69.3711 |

TABLE V
SHADOWS QUALITY WITH DIFFERENT FRAME NUMBERS IN SUB-VIDEOS.

| Number | EPI | Entropy | Contrast |
|---|---|---|---|
| -- | 1.0000 | 8.8025 | 21.0104 |
| 50 | 0.7247 | 6.3759 | 58.5289 |
| 75 | 0.7658 | 6.5247 | 66.8366 |
| 100 | **1.0509** | **4.9927** | 69.3711 |
| 125 | 0.8266 | 6.7042 | 80.0892 |
| 150 | 0.8175 | 6.6799 | **81.7973** |

## V. DISCUSSIONS

### A. Different Noise Distributions

TABLE IV compares the shadow enhancement performance of different noise distributions. From TABLE IV, all different distributions can enhance shadows greatly because the Contrast is increased. The Rayleigh distribution offers a higher Contrast than others, but its EPI and Entropy are inferior to the Gaussian distribution. Thus, to ensure that the shadow shape (i.e., EPI) is not affected meanwhile improving the shadow saliency, the Gaussian distribution is still selected. A more accurate noise modeling of Video-SAR can be considered in the future.

### B. Different Frame Numbers in Sub-Videos

TABLE V compares the shadow enhancement performance of different frame numbers in sub-videos. From TABLE V, the more frames in sub-videos, the higher the shadow contrast, but the EPI and entropy become worse. The number of frames in sub-videos when using SBN-3D-SD is set to the trade-off value 100. This can also ensure that the moving target keeps appearing in the sub-video without leaving the observation area.

## VI. CONCLUSIONS

SBN-3D-SD is proposed to enhance moving target shadows in Video-SAR. It utilizes the sparse property of shadows, the low-rank property of backgrounds, and the Gaussian property of noises to separate shadows from backgrounds and noises. ADMM is used to solve SBN-3D-SD. With it, the moving target shadows are enhanced by three times; the shadow detection accuracy is improved by ∼8%; the shadow tracking accuracy is improved by ∼10%. SBN-3D-SD also has a satisfactory ability of target shadow information retention. It is a powerful pre-processing tool of video/image data for moving target surveillance. It can be applied to video data obtained in many occasions, e.g., traffic control, sensitive target reconnaissance, etc.